\documentclass[letterpaper, 10 pt, conference]{ieeeconf}
\usepackage{cite}
\usepackage{amsmath,amsfonts}
\usepackage{algorithmic}
\usepackage{algorithm}
\usepackage{array}
\usepackage{graphicx}
\usepackage[caption=false,font=footnotesize,labelfont=sf,textfont=sf]{subfig}

\IEEEoverridecommandlockouts                              

\begin{document}
\title{\LARGE \bf
Robust Multi-Robot Global Localization with Unknown Initial Pose based on Neighbor Constraints
}

\author{Yaojie Zhang$^{1,2}$  Haowen Luo$^{1,2}$  Weijun Wang$^{1,2}$  Wei Feng$^{*1,2,3}$
\thanks{This work was supported by the National Natural Science Foundation of China(U20A20283), and the CAS Science and Technology Service Network Plan (STS) - Dongguan Special Project (Grant No. 20211600200062), and the Guangdong Provincial Key Laboratory of Construction Robotics and Intelligent Construction (2022KSYS013), and the Science and Technology Cooperation Special Project of Hubei province and Chinese Academy of Sciences (2023-01-08).}
\thanks{$^{1,2}$Yaojie Zhang, Haowen Luo, and Weijun Wang are with University of Chinese Academy of Sciences, Beijing 100049; the Shenzhen Institute of Advanced Technology, Chinese Academy of Sciences, Shenzhen 518055.  ({\tt\small yj.zhang1@siat.ac.cn})
}%
\thanks{$^{3}$Wei Feng is with the Guangdong Provincial Key Lab of Robotics and Intelligent System and Guangdong Provincial Key Lab of Construction Robotics and Intelligent Manufacturing. ({\tt\small wei.feng@siat.ac.cn})%
}}

\maketitle

\thispagestyle{empty}
\pagestyle{empty}

\begin{abstract}

Multi-robot global localization (MR-GL)  with unknown initial positions in a large scale environment is a challenging task. The key point is the data association between different robots' viewpoints. It also makes  traditional Appearance-based localization methods unusable. Recently, researchers have utilized the object's semantic invariance to generate a semantic graph to address this issue. However, previous works lack robustness and are sensitive to overlap rate of maps, resulting in unpredictable performance in real-world environments. In this paper, we propose a data association algorithm based on neighbor constraints to improve the robustness of the system. We demonstrate the effectiveness of our method on three different datasets, indicating a significant improvement in robustness compared to previous works.

\end{abstract}

\section{INTRODUCTION}

Multi-robot collaborative simultaneous localization and mapping (SLAM) is a promising approach that offers significant advantages over single robot SLAM in various robotic applications, particularly in large-scale environments such as factory automation, search and rescue, surveillance, and intelligent transportation \cite{tian2022kimera}. In the field of multi-robot systems, a fundamental challenge is achieving global localization among individual robots. Only then, robots can share the same coordinate system to collaborate with each other. This task is particularly formidable when the coordinate transformation between the initial poses of the robots is unknown \cite{zhou2006multi}.

The primary objective of MR-GL is to identify common landmark matches between query map and target map. However, correctly identifying corresponding features across different robots is more challenging than in single robot SLAM. This difficulty is primarily due to the significant differences in viewpoints between individual robots. In such conditions, conventional appearance-based methods, such as Bag-of-Words(BoW) \cite{2012Bags}, tend to fail as they are sensitive to changes in viewpoint or lighting.

Recent research has explored the use of semantic invariant characteristics to address the issue of viewpoint changes \cite{liu2019global,gawel2018x,guo2021semantic,ran2021rs}. Previous studies, as documented in \cite{gawel2018x,guo2021semantic}, have demonstrated superior performance compared to appearance-based methods like \cite{arandjelovic2016netvlad,2012Bags} when dealing with large viewpoint changes. These methods employ a semantic label and geometry center of surroundings to generate a semantic graph for each map, which is then used to identify common landmark matches through graph matching techniques. To improve efficiency, the graph descriptor is utilized to extract local graph information of each node, enabling real-time approximate matching.

\begin{figure}[!t]
\centering
\subfloat[Ground Truth]{\includegraphics[width=1.1in]{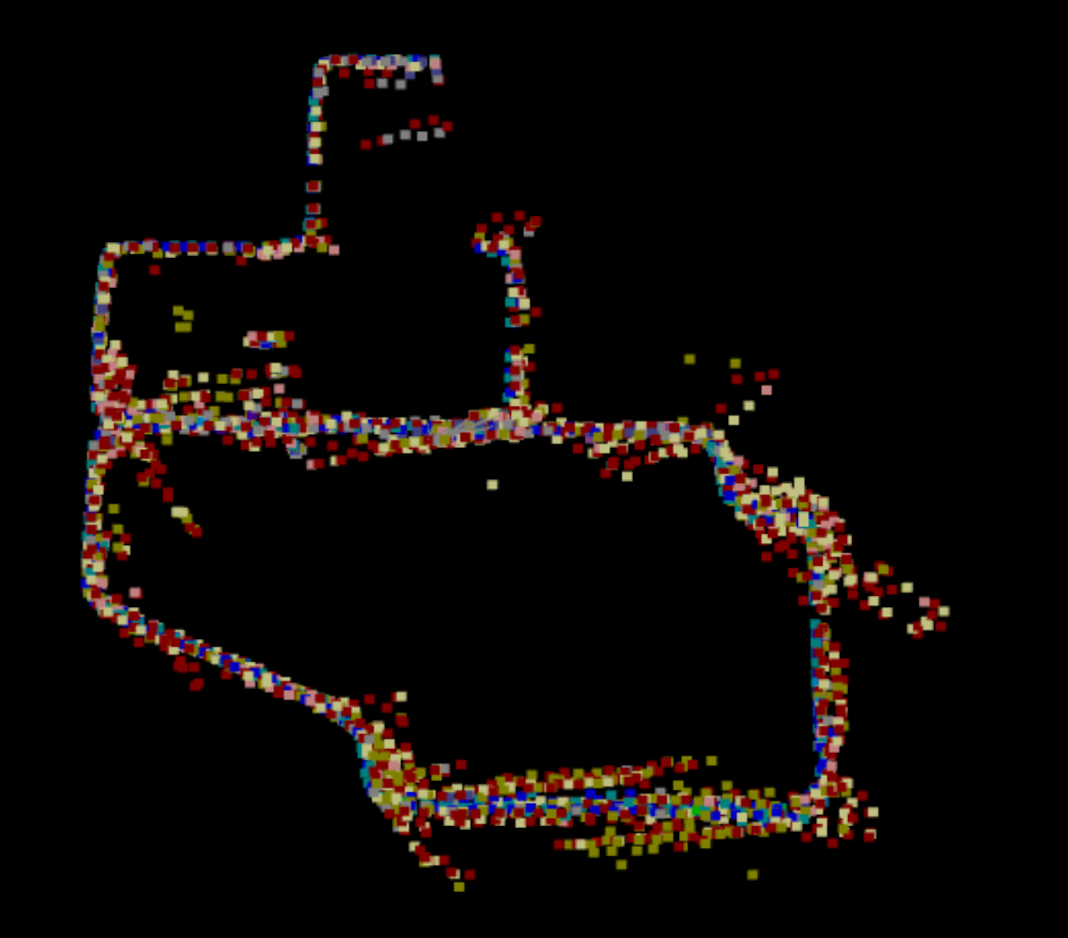}%
\label{fig_1_a}}
\hfil
\subfloat[Previous]{\includegraphics[width=1.1in]{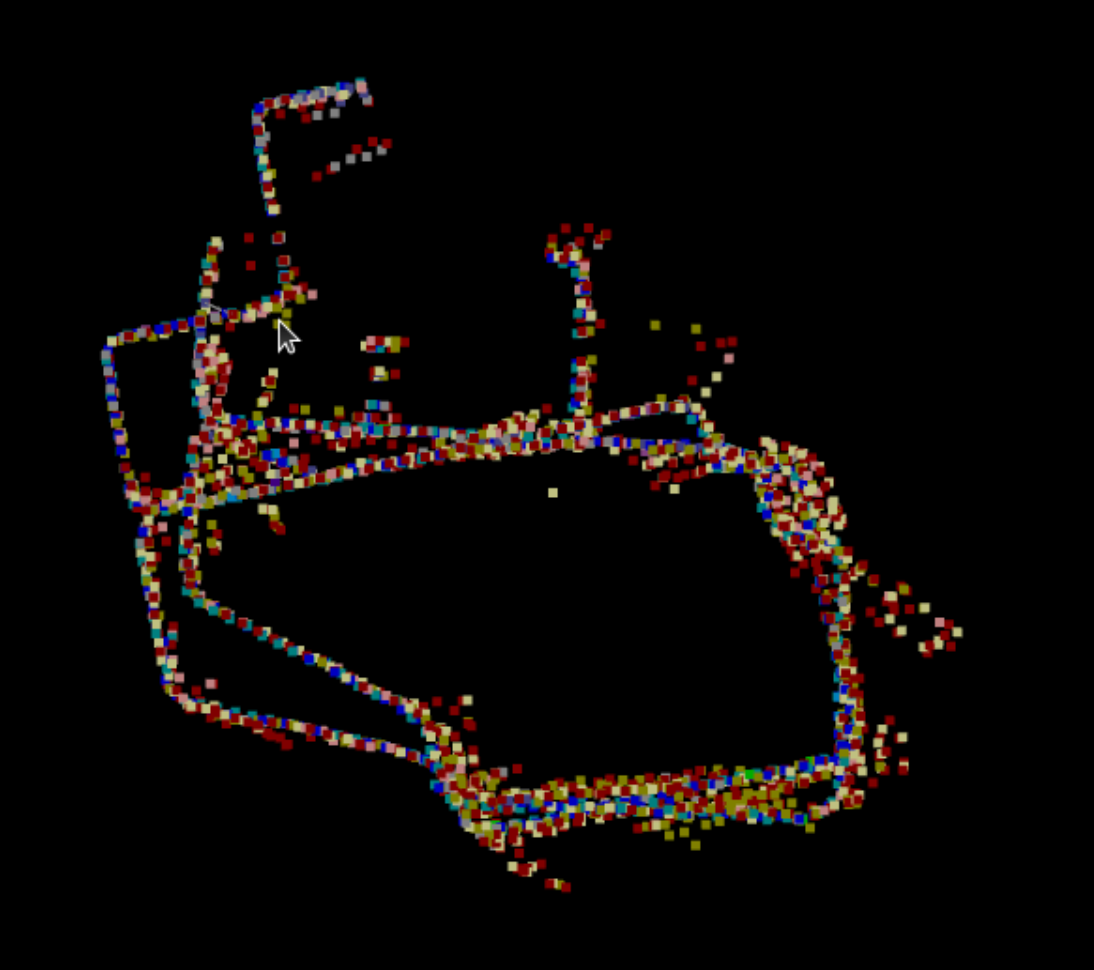}%
\label{fig_1_b}}
\hfil
\subfloat[Ours]{\includegraphics[width=1.1in]{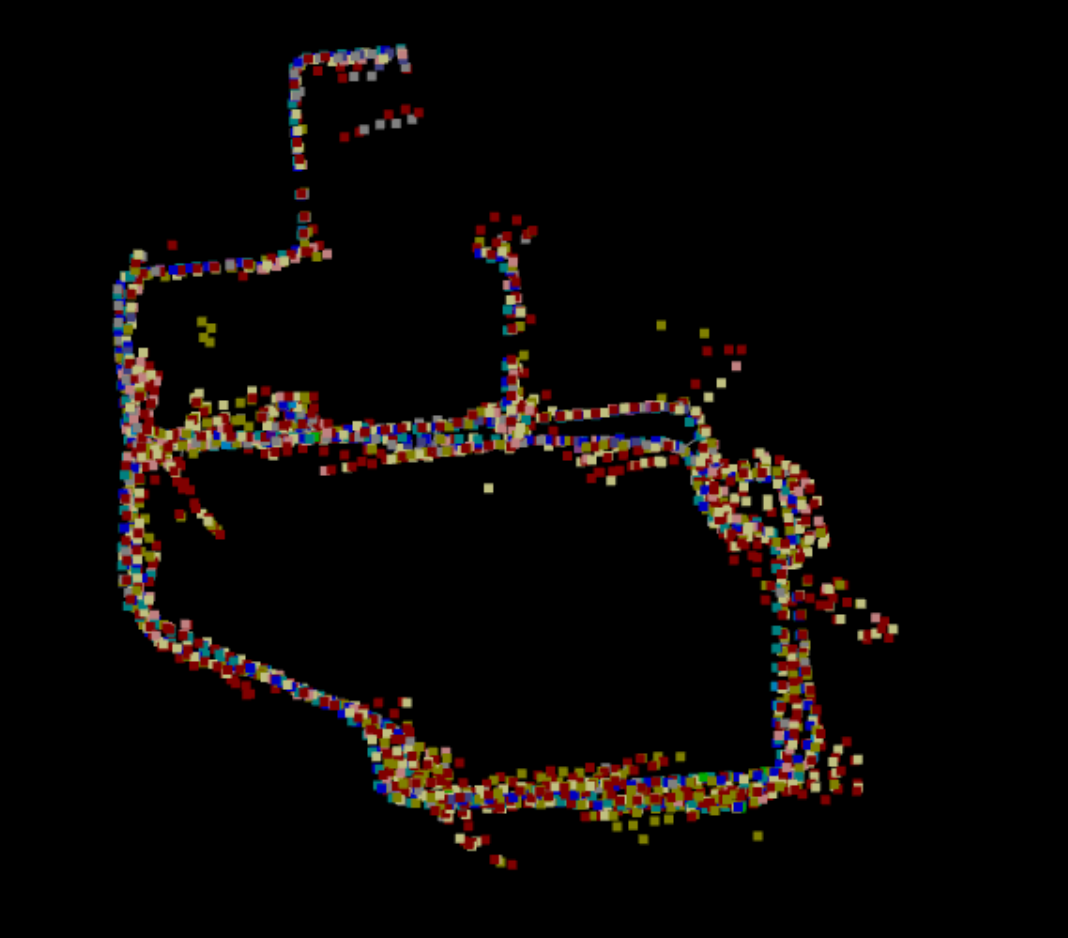}%
\label{fig_1_c}}
\caption{\textbf{An example of a challenging task.} In this task, each graph has about 1300 nodes and exist lots of repetitive scenarios. The figure shows the worst localization condition of previous method as well as ours. With the previous method, the result has 9\% probability failure rate (Translation error over 20m).}
\label{fig_1}
\end{figure}

In the context of map merging, it is typically assumed that the two maps being integrated have some degree of overlap. However, prior researches \cite{gawel2018x,guo2021semantic} have highlighted the limited robustness of this approach when the overlap rate is too low. The primary drawback of previous graph descriptor method is that they only utilize the semantic information of nodes, thus wasting the neighbor or distance relationship between matches. To address this issue, this paper proposes an algorithm that leverages neighbor constraints to reject outlier matches. Utilizing this neighbor constraints, we can significantly improve the robustness of the system. For an example shown in Fig.\ref{fig_1}, the previous method might produce a bad localization result while our method can avoid.

Our main contributes are:
\begin{itemize}
    \item We firstly introduce a preliminary outlier rejection procedure into the pipeline and propose a simple yet fundamental reject algorithm based on neighbor constraints.

\item  We resolve the unpredictable result issue under low overlap rate condition.

\item We conduct experiments on different datasets to fully and fairly evaluate the performance. Experiment results demonstrate that our approach outperforms previous works in multiple aspects, including precision, robustness, and processing time.
\end{itemize}

\begin{figure*}[!t]
\centering
\includegraphics[width=7in]{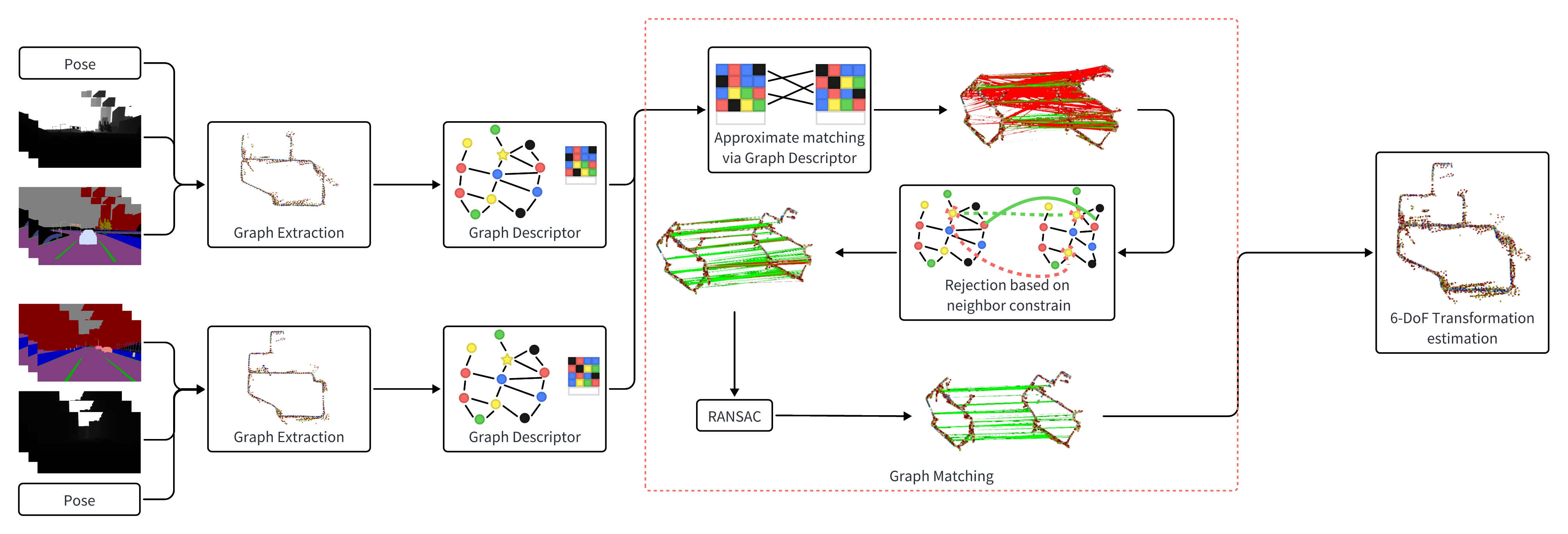}
\caption{\textbf{Our approach overall architecture.} The 3D semantic graph for each robot is first built from semantic frames, depth frames, and poses. Then, the descriptor of each node is extracted to achieve approximate graph matching. Next, the two graphs are matched by comparing the descriptors across the graphs. To be more specific, ours method utilize the neighbor constraints to make a preliminary rejection. Finally, two robots can achieve global localization using the corrected matched correspondences.}
\label{arch}
\end{figure*}

\section{Related Work}
In this section, we review previous approaches in multi-robot global localization. And these approaches can be divided into traditional appearance-based approaches and Graph-based approaches.
\subsection{Traditional Appearance-based Approaches}
A common appearance-based localization method is Bag-of-Words(Bow), giving reliable performance under small viewpoint difference conditions. Localization systems that use the Bag-of-Words approach include \cite{bay2006surf,rublee2011orb,angeli2008incremental,cummins2011appearance,filliat2007visual}. However, these methods are sensitive to large viewpoint difference and light changing. Several place recognition approaches based on convolutional neural network (CNN) architecture were proposed to overcome the viewpoint changing problem \cite{arandjelovic2016netvlad,chen2017deep,sunderhauf2015place,garg2018lost}. Learning from a large number of labeled images taken from different time and location, the CNN based approach provides a more robust performance to viewpoint difference as well as light changing. However, these approaches require large amounts of training data and are computationally expensive. And they only focused on place recognition, the localization is not considered.

\subsection{Graph-Based Approaches}
Graph-based methods formulate the global localization problem as a graph registration problem. These methods extract correspondences between nodes across graphs to determine associations between different graphs, and then calculate the relative pose between graphs. 

Several recent methods have employed semantics to generate labels \cite{li2019semantic,finman2015toward,stumm2016robust,gawel2018x,guo2021semantic}. Utilizing semantic invariant characteristics, these approaches show great potential for improving the reliability of localization systems in scenarios with large viewpoint differences. However, the exact Graph matching is a NP-complete problem \cite{mckay2014practical} which takes exponential time to solve. To balance prediction accuracy and algorithm real-time performance, a series of graph descriptors have been proposed for approximate matching. In \cite{gawel2018x}, it provides a complete multi-robot global localization system using random walk descriptor. In \cite{guo2021semantic}, the author embedding the random walk descriptor into a histogram based descriptor to make a faster performance, and makes the system can be deployed in large environment.

All these descriptor based methods can only find approximate solution for the graph matching problem. To ensure accurate localization performance, the RANSAC \cite{fischler1981random} algorithm is used to reject outliers. However, we find the use of RANSAC can become a bottleneck for both robustness and real-time performance. 

To address this issue, we firstly introduce a preliminary outlier rejection procedure into the pipeline of \cite{gawel2018x} and \cite{guo2021semantic}. In general 3D point cloud registration (PCR) field, numerous outlier rejection techniques are available  \cite{zhang20233d, yang2020teaser, lusk2021clipper, bai2021pointdsc, chen2022sc2, yang2015go}. Many of these approaches rely on spectral matching methods  \cite{leordeanu2005spectral} as a critical component. Therefore, rather than exhaustively testing all these methods, we propose a simple yet fundamental reject algorithm based on neighbor constraints which is a variation of \cite{leordeanu2005spectral} to verify the effectiveness of our proposed procedure.

\section{Method}
In this section, we present our system for global localization. The architecture of our frame work is shown in Fig.\ref{arch}. Our method is mainly based on the previous works \cite{gawel2018x} and \cite{guo2021semantic}. Thus, we will only briefly introduce the common part with previous works. Our algorithm is mainly described in the section C.
\subsection{Graph Extraction}
In accordance with prior researches \cite{gawel2018x,guo2021semantic}, this step employs semantic and depth images to extract graph vertices. These images can be obtained from a mono image using the Deep Neural Network method described in \cite{zhu2019improving,lee2019big}. Initially, we use Seed Filling method \cite{smith1979tint} on the semantic image to segment objects and obtain the object's 2D center position as well as vertex label $l_j$. Subsequently, we combine the depth information to obtain the 3D location $p_j$ of the object. And we ultimately obtain vertices $v_j=\{p_j,l_j\}$. By computing the Euclidean distance between any two vertices, we treat vertices as connected if the distance is less than a threshold. With all vertices and edges, we can obtain a Graph for each robot, denoted as $G_q$ and $G_t$.

In addition to previous works, we also maintain distance matrices for each Graph to exhaust the geometry information. Under this condition, we can treat $G_q$ and $G_t$ as weighted Graphs. Our system is based on weighted Graphs, which will be further elaborated in Section C.

\subsection{Graph descriptor and matching}
Upon creating $G_q$ and $G_t$, the task of finding common landmark matches is reduced to a sub-graph matching issue between two graphs. However, sub-graph matching is an NP-Complete problem \cite{mckay2014practical}, and computation time increases exponentially with the number of vertices. To perform real-time global localization, we only compute the similarity score between corresponding graph descriptors to identify approximate matches.

The graph descriptor is a vector which embeds each vertex's information as well as its surrounding graph structure. Various graph descriptors exist, and this paper primarily tests Fast \cite{stumm2016robust}, Random Walk \cite{gawel2018x}, and Histogram-based descriptors \cite{guo2021semantic}. 

Each vertex corresponds to a descriptor. We then compute the similarity score between descriptors $D_j$ and $D_i$ using below cosine similarity, and matches with scores higher than a threshold are chosen as our original matches.
\begin{equation}
    similarity_{ij} = \frac{D_i\cdot D_j}{||D_i||\ ||D_j||}\tag{1}
\end{equation}

\subsection{Rejection}

\begin{algorithm}[!t]
\caption{Rejection based on neighbor constraints.}\label{alg:alg1}
\begin{algorithmic}
\STATE
\STATE {\textsc{Input:\ }}
\STATE \hspace{0.5cm} $\mathbf{M_0}$ \ Original\ Matches\ $\mathbf{(m\times2)}$,\\
\STATE \hspace{0.5cm} $\mathbf{G_q}$\ and\ $\mathbf{G_t}$:\ weighted\ graph\\
\STATE {\textsc{Output:\ }} 
\STATE \hspace{0.5cm} $\mathbf{M_{out}}$:\ final Matches\ $\mathbf{(n\times2)}$\\
\STATE Initialize\ $\mathbf{M^{m\times m}}$\ to\ all\ zeros\\
\STATE {Initialize}\ $\mathbf{V^{{1\times m}}}$\ to\ all\ zeros\\
\STATE \textit{Step 1: Generate neighbor constraint matrix} $\mathbf{M}$.
\STATE \textbf{for} $M_{ij}$ in $\mathbf{M}$:\\
\STATE \hspace{0.5cm} $G_q^i \leftarrow $Index$\ \mathbf{M_0}[i][0]\ $of$\ \mathbf{G_q}$\
\STATE \hspace{0.5cm} $G_q^j \leftarrow $Index$\ \mathbf{M_0}[j][0]\ $of$\ \mathbf{G_q}$\
\STATE \hspace{0.5cm} $G_t^i \leftarrow $Index$\ \mathbf{M_0}[i][1]\ $of$\ \mathbf{G_t}$\
\STATE \hspace{0.5cm} $G_t^j \leftarrow $Index$\ \mathbf{M_0}[j][1]\ $of$\ \mathbf{G_t}$\
\STATE \hspace{0.5cm} \textbf{if} $|{D}(G^i_q,G^j_q)-{D}(G^i_t,G^j_t)|<threshold$:\\
\STATE \hspace{1cm}		$M_{ij} \leftarrow 1$\\
\STATE \hspace{0.5cm}	\textbf{end\ if}\\
\STATE \textbf{end\ for}\\
\STATE \textit{Step 2: Reject bad matches}.
\STATE $\mathbf{V}$ $\leftarrow$ add\ all\ rows\ of\ $\mathbf{M}$\\
\STATE i $\leftarrow$ $argmin(V_i\ in\ \mathbf{V})$\\
\STATE \textbf{while}\ ($V_{i} < 0.5 * rows\ of\ \mathbf{M_{0}}$):\\
\STATE \hspace{0.5cm} $\mathbf{V} \leftarrow$ $\mathbf{V} -$ $\mathbf{M}[i,:]$ $\COMMENT{Subtraction\ i_{th}\ row\ of\ \mathbf{M}}$\\ 
\STATE \hspace{0.5cm} $V_{i} \leftarrow m\ \COMMENT{Update\ V_i\ a\ large\ number}$\\
\STATE \hspace{0.5cm} $delete\ i_{th}$\ row\ of\ $\mathbf{M_0}$\\
\STATE \hspace{0.5cm}	i $\leftarrow$ $argmin(V_i\ in\ \mathbf{V})$\\
\STATE \textbf{end while}\\
\STATE $\mathbf{M_{out}} \leftarrow \mathbf{M_0}$\\
\STATE return $\mathbf{M_{out}}$

\end{algorithmic}
\label{alg1}
\end{algorithm}

After obtaining vertex matches between $G_q$ and $G_t$, numerous mismatches are observed due to object recurrence in large environments. For instance, in urban cities, many objects such as trees, buildings, roads, and poles may have high similarity scores even though in different areas. Previous work directly applies RANSAC to the original matches. However, the RANSAC will have low precision performance and efficiency due to numerous incorrect matches. Therefore, we first apply our neighbor constraints algorithm to reject most incorrect matches and obtain a set of better intermediate matches. Subsequently, we use the RANSAC method to obtain the final matches.

The core idea of our neighbor constraints algorithm is that the vertices in $G_q$ are neighbors, and the matched vertices in $G_t$ should also be neighbors. Otherwise, there exists at least one wrong match between two matches. For the match whose neighbor relationship is the worst, we can treat it as a wrong match and reject it. The algorithm is depicted as follows: first, a neighbor matrix, $M$, is generated from adjacency matrix of weighted graph $G_q$ and $G_t$. This neighbor matrix's form is similar to the adjacency matrix of a graph, but depicts the neighbor relationship between matches instead of vertices. The value of $M_{ij}$ is 1 if vertices from two matches have the same neighbor relationship in both $G_q$ and $G_t$; otherwise, it is set to 0. The mathematical expression is as follows:

\begin{equation}
M_{ij} = \left\{
\begin{aligned}
1, &\ \textsc{Nei}(G^i_q,G^j_q)=\textsc{Nei}(G^i_t,G^j_t) \\
0, &\ \textsc{Nei}(G^i_q,G^j_q)\neq\textsc{Nei}(G^i_t,G^j_t)
\end{aligned}
\tag{2}
\right.
\end{equation}

\begin{equation}
\textsc{Nei(A,B)} = \left\{
\begin{aligned}
True, &\ Distance(A,B) < threshold\\
False, &\ Otherwise
\end{aligned}
\tag{3}
\right.
\end{equation}
Where the $G^i_q$ represents the vertex of $i_{th}$ matches corresponding to weighted graph $G_q$, and it contains the 3D position of the object. The same expression in regard to $G^j_q$, $G^i_t$ and $G^j_t$.

Subsequently, we reject the worst matches that has the worst neighbor relationship among all matches. And repeat this step until there is no bad match, finally get $n$ matches which much fewer than original matches. Our detailed algorithm is shown as Alg.\ref{alg1}. The bad match is defined as:

\begin{equation}
    \frac{\#\ of\ correct\ Neighbor\ relationship}{\#\ of\ total\ Matches} > \eta .
\notag
\end{equation}

Our proposed rejection algorithm has several properties, which we also demonstrate in the experiment section. Firstly, the algorithm always rejects the worst match with the least neighbor score among other matches. Secondly, it can avoid most incorrect matches, allowing us to obtain a more robust RANSAC solution closer to optimal. Thirdly, this method can reject some special occlusions that RANSAC cannot.

\subsection{Pose Estimation}
Upon obtaining the final matches, we can get correspond vertices or point clouds $p_q$ and $p_t$ from $G_q$ and $G_t$ for registration. Then, we can use the ICP algorithm \cite{rusinkiewicz2001efficient} to get the Rotation matrix $R$ and translation vector $T$ by minimizing the following squared error metric:
\begin{equation}
E(R,t) = \sum_{i=1}^{n} W_i||Rp^i_t+T-p^i_q||^2 \tag{4}
\end{equation}
The points in $p_q$ are denoted as $p^i_q$ and the points in $p_t$ are denoted as $p^i_t$, where $i$ ranges from 1 to $n$. $W_i$ is the weight factor, which is related to the corresponding objects’ shape.

\section{Experiments}
\begin{figure}[!t]
\centering
\captionsetup[subfloat]{labelsep=none,format=plain,labelformat=empty}
\subfloat{\includegraphics[width=1in]{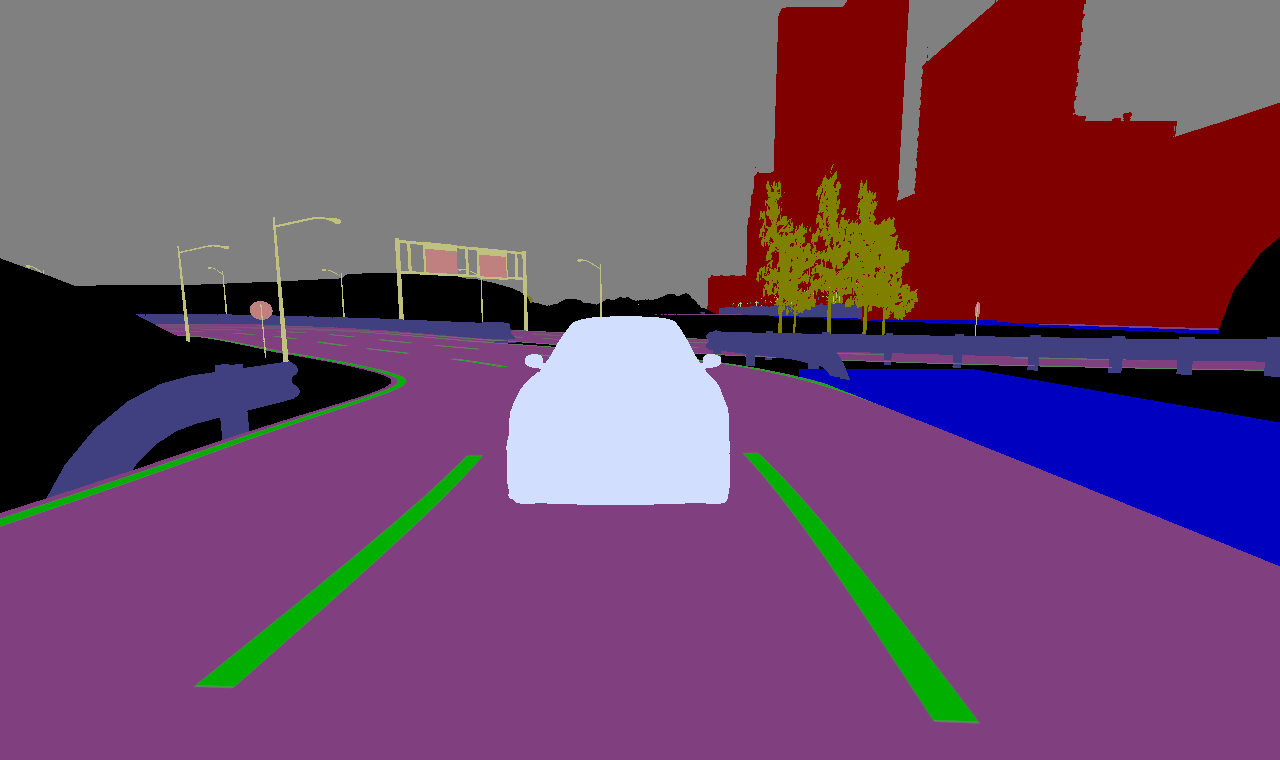}}
\ 
\subfloat{\includegraphics[width=1in]{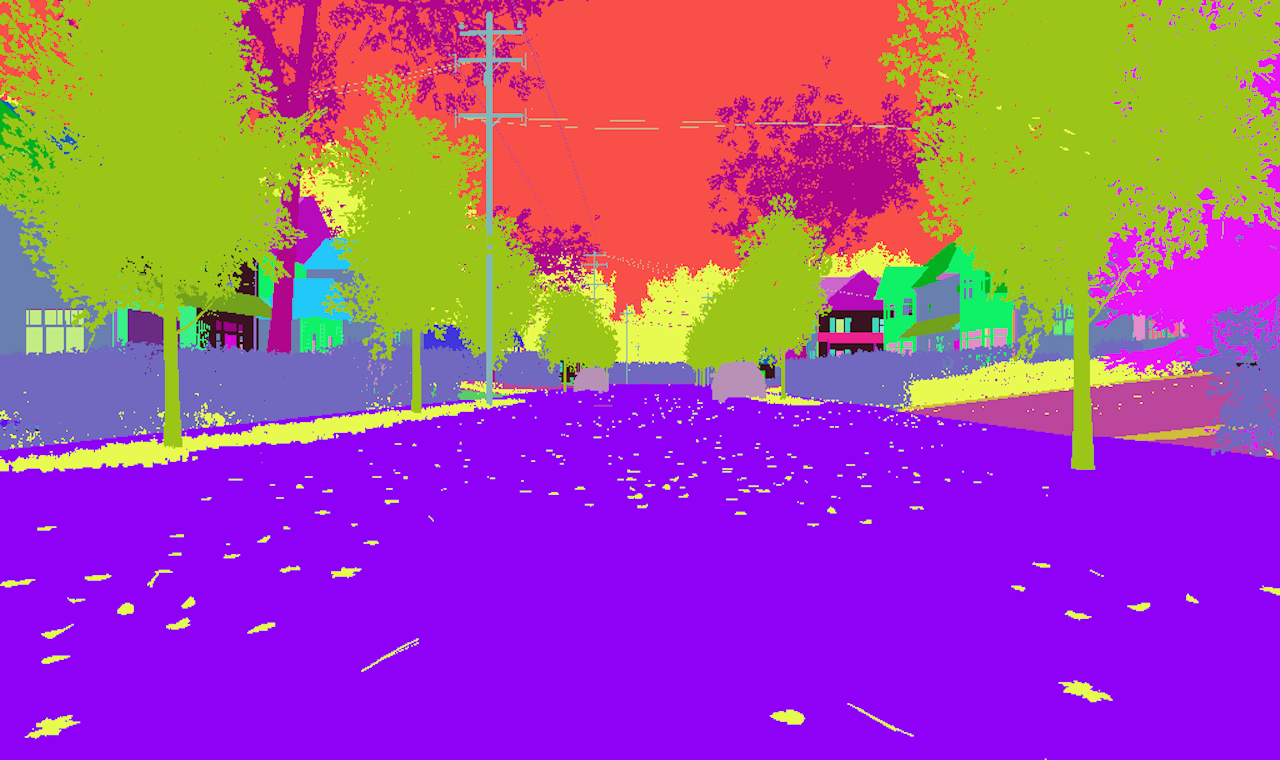}}
\ 
\subfloat{\includegraphics[width=1in]{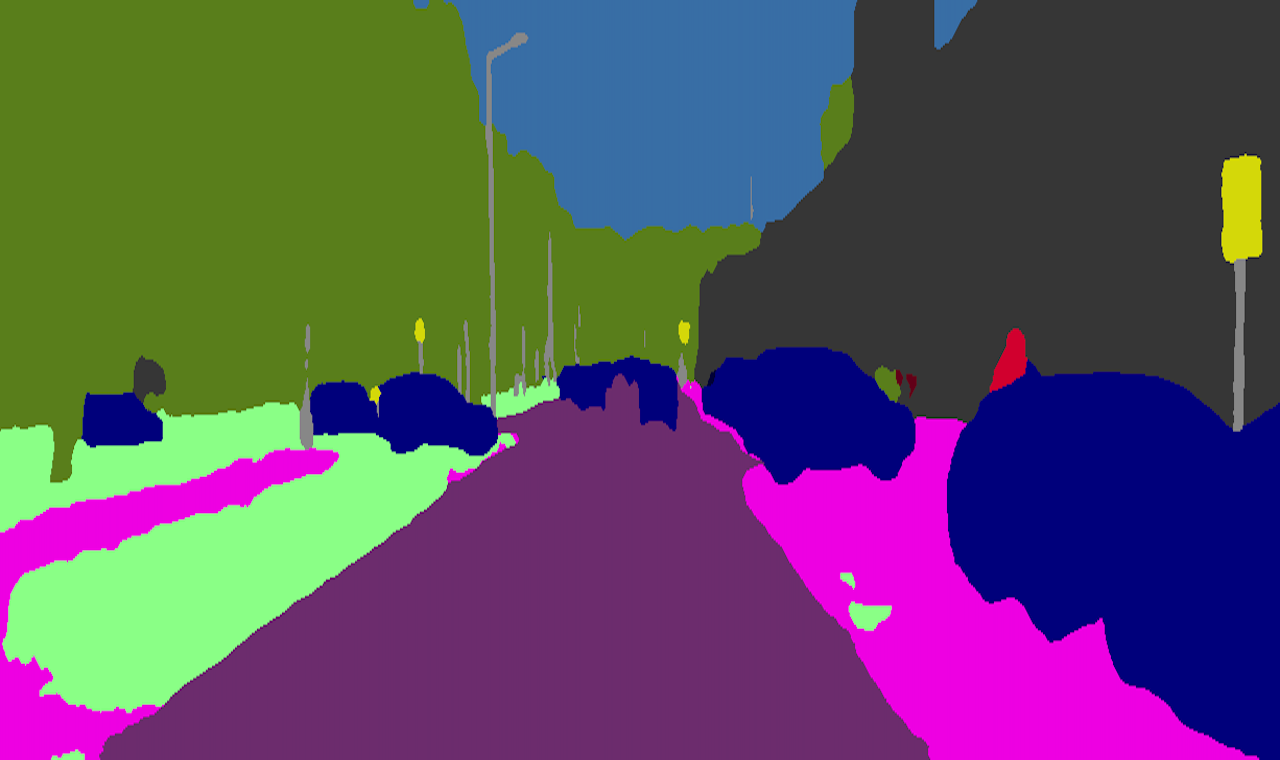}}
\hfill
\subfloat[(a) SYNTHIA]{\includegraphics[width=1in]{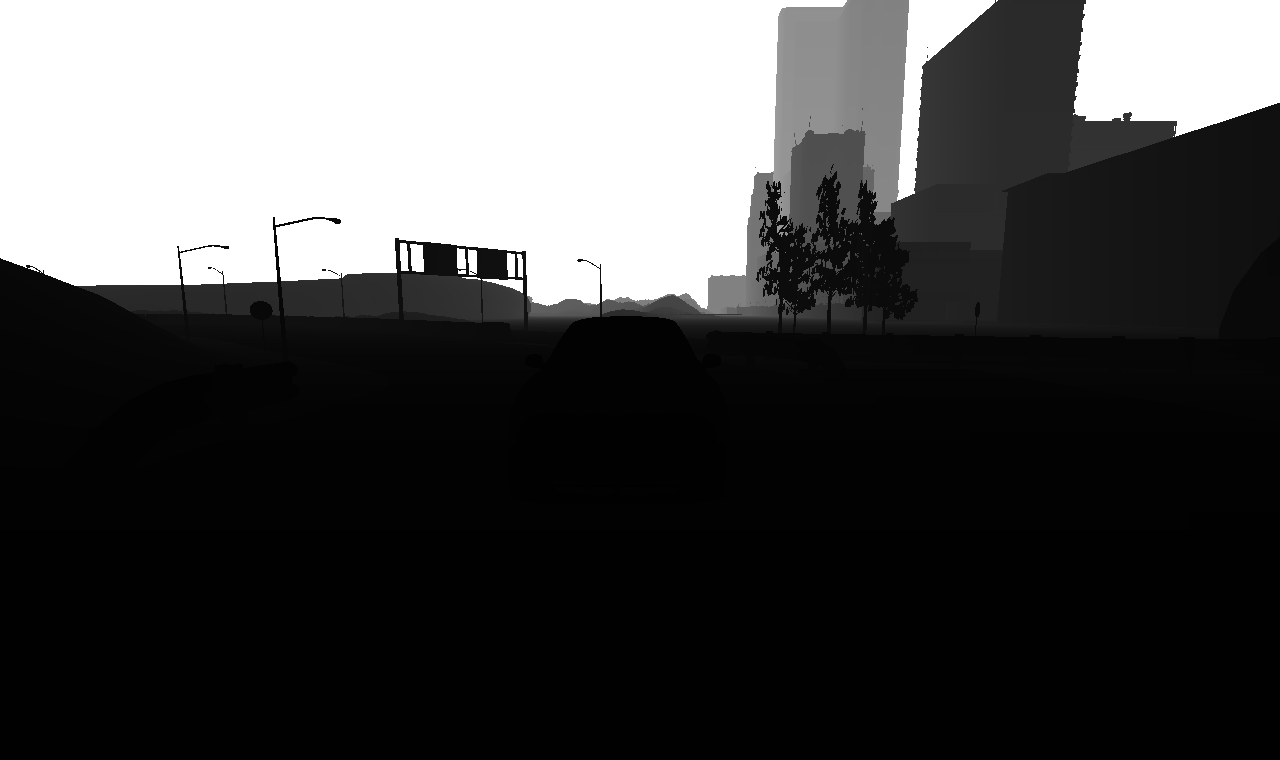}%
\label{fig_3_a}}
\ 
\subfloat[(b) Airsim]{\includegraphics[width=1in]{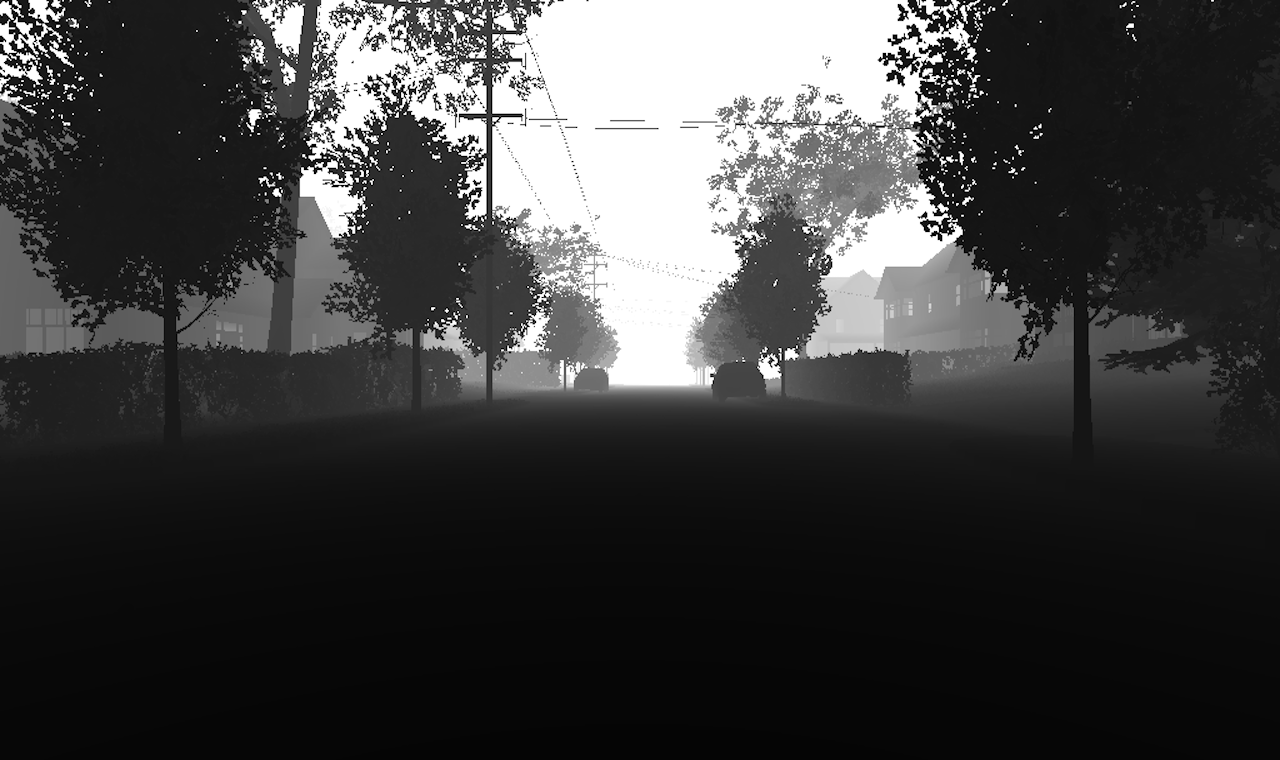}%
\label{fig_3_b}}
\ 
\subfloat[(c) KITTI]{\includegraphics[width=1in]{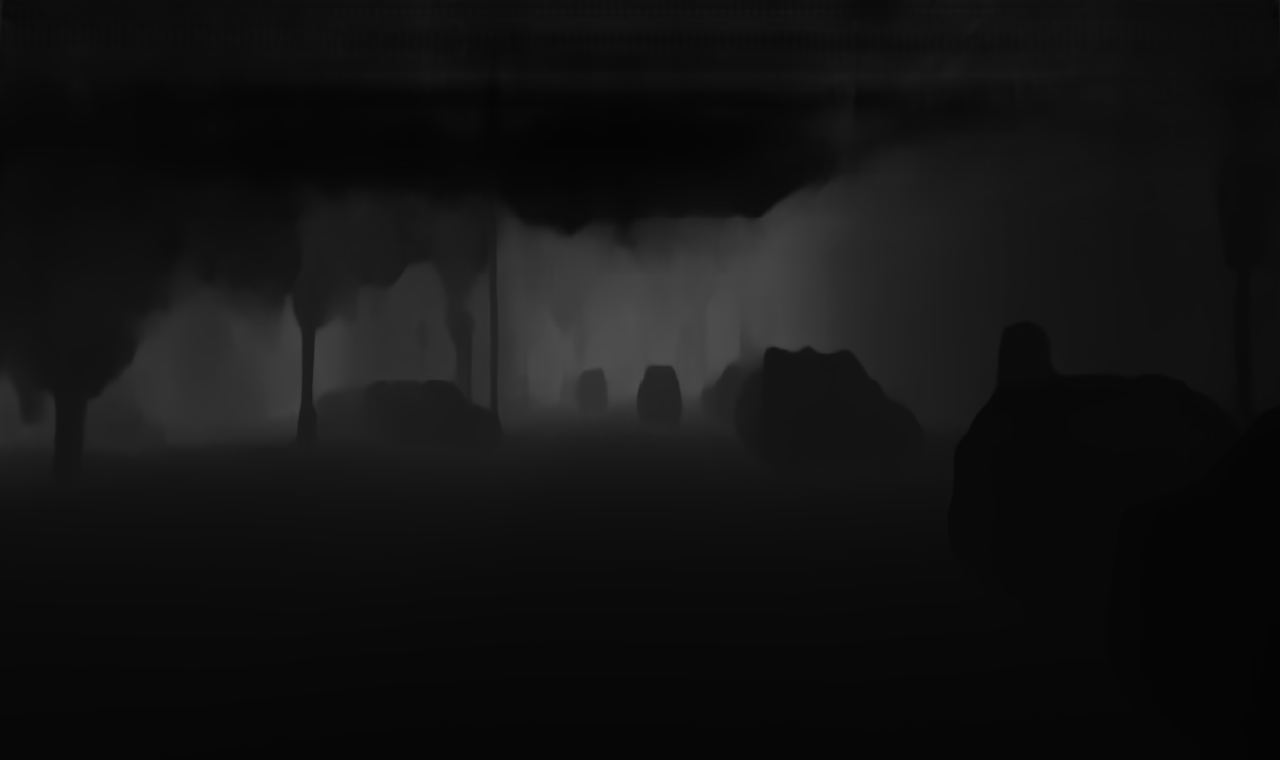}%
\label{fig_3_c}}
\caption{\textbf{Sample images from three datasets used in the experiments.} (top) Semantic segmentation, (bottom) Depth image. SYNTHIA with perfect semantic segmentation and depth image, Airsim with perfect semantic segmentation and depth image, KITTI with LabelRelatx \cite{zhu2019improving} semantic segmentation and BTS \cite{lee2019big} depth image.}
\label{fig_3}
\end{figure}

We conduct three experiments on three different datasets with different segmentation situations shown in Fig.\ref{fig_3} to validate the effectiveness of our method. Firstly, we make quantitative comparisons with the previous work \cite{guo2021semantic} on the Airsim dataset \cite{shah2018airsim}, mainly focusing on robustness, precision, and runtime performance. Secondly, on the SYNTHIA dataset \cite{ros2016synthia},we show that our method is more robust to the overlap rate of the map. Lastly, we conduct an experiment on the real-world KITTI dataset \cite{geiger2013vision}. We select an environment with few object types and poor segmentation result to validate our method's performance under challenging conditions. All experiments are performed on a Intel Core i7-9700 CPU @3.0GHz with 8GB RAM.

\begin{figure}[!b]
\centering
\subfloat[Translation error]{\includegraphics[width=1.5in]{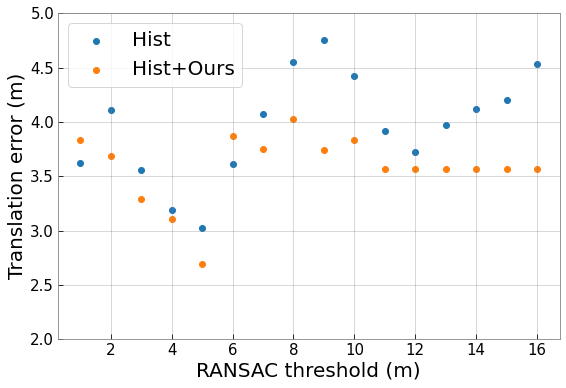}%
\label{fig_4_a}}
\hfil
\subfloat[Distributions]{\includegraphics[width=1.5in]{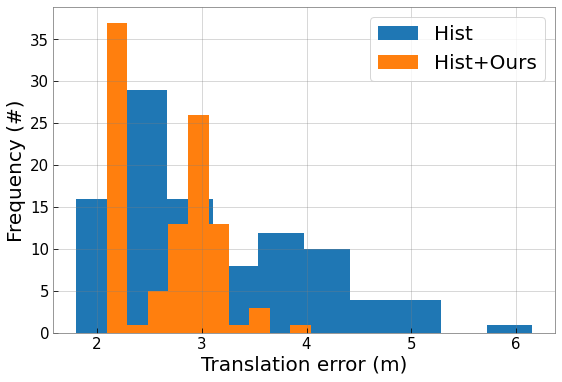}%
\label{fig_4_b}}
\caption{\textbf{Detailed translation error comparisons.} The blue points/bars are previous approach's results and orange is for ours.(a) The RANSAC threshold represents the accept deviation for RANSAC. Translation error represents in Euclidean distance(m). Each point is an average value of 100 times results. (b) Distributions on RANSAC threshold 5m.}
\label{fig_4}
\end{figure}

\subsection{Airsim}
\subsubsection{Dataset and Implementation Details}
We use the same Airsim dataset provided by \cite{guo2021semantic} to reproduce the experiment, ensuring a fair comparison with previous work. The dataset simulates an urban block using Airsim, including two car's trajectories. The Airsim dataset provides three types of data: RGB images, depth maps, and semantic maps. The average travel distance of the cars is 420 meters, and the two cars' directions are opposite at overlapping parts (over 200m), indicating an extremely large viewpoint change.

To demonstrate that our method is suitable for different kinds of descriptors, we conducted experiments using three different graph descriptors: Histogram, Fast and Random walk descriptor. We maintained all other system parameters exactly the same and only added our algorithm to make a comparison. We evaluated the matching performance based on the translation error, rotation error, and processing time. We also changed the RANSAC threshold to show that our algorithm is suitable for different acceptable deviation matches and can improve matching performance. Since the RANSAC method is a random algorithm, each result is an average value calculated over 100 iterations.

\begin{table*}[!t]
\caption{The quantitative comparisons of previous and ours method with different descriptors\label{tab:table1}}
\centering
\begin{tabular}{c|ccc|ccc|cc}
\hline  Descriptor& Final &Recall &Precision & Ours & RANSAC &Total  &Translation& Rotation Error\\  Type &Matches &Rate(\%) &Rate(\%)  &Time(ms) &Time(ms) &Time(ms) &Error(m) &Error(degree)\\
\hline 
Histogram&67.84$\pm$1.55&85.2&77.3&-&19.2&19.2&3.02$\pm$0.91 &0.41$\pm$0.32\\ Hist+Ours&$\mathbf{72.67\pm0.78}$&$\mathbf{91.4}$&$\mathbf{87.9}$&1.2&7.9&$\mathbf{9.1}$&$\mathbf{2.69\pm0.43}$&$\mathbf{0.40\pm0.22}$\\
\hline 
Fast\ Descriptor &${24.84\pm3.47}$ &65.4 &$\mathbf{73.3}$ &- &18.4 &18.4 &5.72$\pm$6.53 &3.88$\pm$5.71\\  Fast+Ours &$\mathbf{36.69\pm1.34}$ &$\mathbf{85.0}$ &65.7 &1.0 &$6.0$ &$\mathbf{7.0}$ &$\mathbf{4.33\pm1.31}$ &$\mathbf{0.85\pm0.28}$\\
\hline
Random Walk&62.90$\pm$1.91&66.3&83.0&-&18.3&18.3&2.76$\pm$0.94 &0.61$\pm$0.33\\
Random Walk+Ours&$\mathbf{68.13\pm0.61}$&$\mathbf{71.3}$&$\mathbf{91.2}$&0.7&11.7&$\mathbf{12.4}$&$\mathbf{2.44\pm0.49}$ &$\mathbf{0.61\pm0.15}$\\
\hline
\end{tabular}
\end{table*}

\begin{figure*}[!t]
\centering
\subfloat[Histogram vs Ours]{\includegraphics[width=2in]{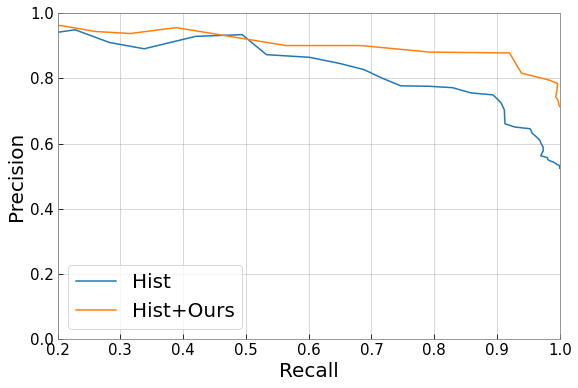}%
\label{fig_5_a}}
\hfil
\subfloat[Fast vs Ours]{\includegraphics[width=2in]{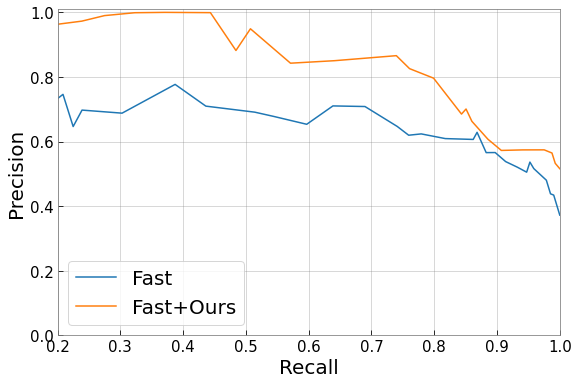}%
\label{fig_5_b}}
\hfil
\subfloat[Random Walk vs Ours]{\includegraphics[width=2in]{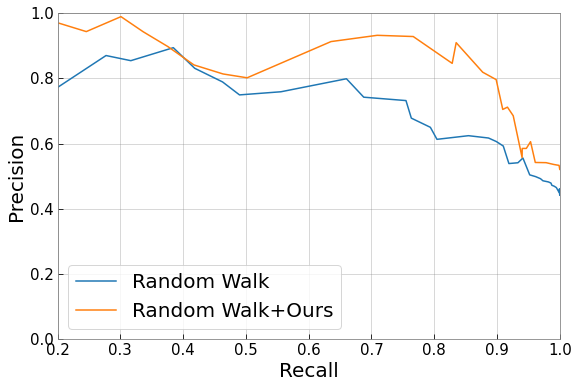}%
\label{fig_5_c}}
\caption{\textbf{PR curve about predicted matches.} The PR curve indicates the ability of finding correct matches. The precision=1 means all predicted matches are good matches. The recall=1 means all good matches are found.}
\label{fig_5}
\end{figure*}

\subsubsection{Experimental Results and analysis}
We present the results of the translation error comparison between our method and the original method in Fig.\ref{fig_4}. Our method outperforms the original method at most different RANSAC thresholds. The Table \ref{tab:table1} provides more detailed comparison of different descriptors when the RANSAC threshold is 5m. Firstly, our method can find more good matches than the original method, even though our method passes on much fewer matches to the RANSAC step. This indicates that our result is closer to the optimal RANSAC result, and our algorithm always reject wrong matches. Secondly, our processing time is about 2 times faster than the original method, despite the introduction of an extra step to the system. This is mainly because RANSAC processing time is saved immensely with much fewer matches. Thirdly, our translation error and rotation error are lower with smaller deviation. We can also observe that our method makes a more significant improvement on the Fast Descriptor. This is mainly due to the fact that Fast Descriptor can find a much lower good matches rate (10\%), and RANSAC is harder to find a good solution under this condition. While our method can help RANSAC make a marginal improvement by rejecting most wrong matches.

To better illustrate the performance of our algorithm, we plotted the PR curve by changing the RANSAC threshold in Fig.\ref{fig_5}. We can divide the predicted matches into inliers and outliers according to the distance threshold ($T_p$=10m). The recall and precision were defined as follows:
$$
    Recall = \frac{\#\ inliers_{\ (predicted)}}{\#\ inliers_{\ (total)}},
$$
$$
    Precision = \frac{\#\ inliers_{\ (predicted)}}{\#\ matches_{\ (predicted)}}
$$
In general, the PR curve of our method are above the previous method for all three descriptors. It indicates that our method always find more accurate matches at the same recall rate and suits well for different kind of descriptors.

\subsection{SYNTHIA}
\subsubsection{Dataset and Implementation Details}
In this experiment, we utilize the publicly available SYNTHIA dataset \cite{ros2016synthia}. The SYNTHIA dataset consists of numerous sequences of simulated sensor data collected from a car navigating through diverse dynamic environments and under varying conditions such as weather and daytime. The sensor data provides RGB, depth, and pixel-wise semantic classification for eight cameras, with two cameras always facing forward, left, backward, and right respectively. To simulate viewpoint variation, we associated data from the forward view with that of the backward view. The semantic classification entails 13 distinct semantic classes that are labeled in a class-wise manner. Furthermore, dynamic objects such as pedestrians and cars are also labeled instance-wise. In this section, we specifically focused on sequence 4, which features a town-like environment and covers a total traveled distance of 970 meters.

The main objective of our experiment is to investigate how the degree of overlapping influences the global localization performance. We changed the overlap rate of trajectories by selecting frames from the forward and backward view. We then compare the global localization performance between our method and the original method using the Histogram based descriptor. The performance is evaluated by translation error and repeated 100 times to obtain a box plot.

\begin{figure}[!b]
\centering
\includegraphics[width=3in]{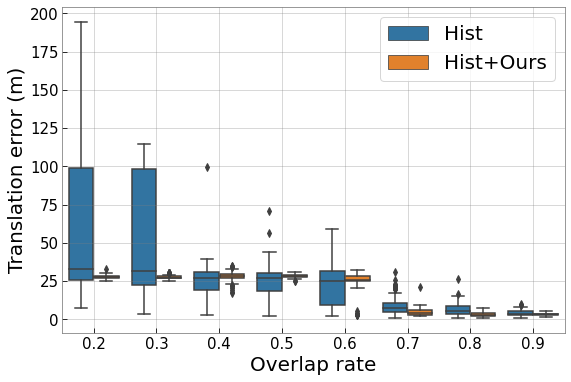}
\caption{\textbf{The influence of different degree of overlap.} Each box plot indicates the distribution of 100 results. The wider of the box, the more unpredictable result of the system. This figure shows that our method can narrow this unpredictable results and have a more robust result.}
\label{fig_6}
\end{figure}

\begin{table}[!t]
\caption{The quantitative results of SYNTHIA dataset\label{tab:table2}}
\centering
\begin{tabular}{c|c|ccc}
\hline  Descriptor &Processing  &Translation& Rotation Error\\  Type &Time(ms) &Error(m) &Error(degree)\\
\hline
Histogram &77.3&8.07$\pm$2.02 &2.51$\pm$0.68\\
Hist+Ours & $\mathbf{47.9}$&$\mathbf{6.83\pm1.04}$&$\mathbf{2.07\pm0.41}$ \\
\hline
Fast\ Descriptor &72.7 &2.57$\pm$1.70 &3.92$\pm$3.11\\  Fast+Ours &$\mathbf{48.4}$ &$\mathbf{1.78\pm1.12}$ &$\mathbf{1.23\pm1.12}$\\
\hline
Random Walk & 73.9 & 13.56$\pm$5.51 &3.83$\pm$3.58\\
Random Walk+Ours&$\mathbf{45.6}$&$\mathbf{7.80\pm2.11}$ &$\mathbf{1.54\pm0.86}$\\
\hline
\end{tabular}
\end{table}

\subsubsection{Experimental Results and analysis}
The results of our investigation into the influence of overlapping are presented in Fig.\ref{fig_6}. It is obvious that our method is more stable than previous method across all overlap rates, particularly when the overlap rate is low. The figure illustrates that the previous method lacks robustness to overlapping and is unpredictable when the overlap rate is below 0.3. This is primarily due to the low rate of good matches, which makes it difficult for the previous method to find a reasonable model under these conditions. In contrast, our method can reject most of the bad matches, ensuring that the overlap rate does not affect the stability of our results. And we also present the quantitative results of whole maps in table \ref{tab:table2}. It is obvious that our method have better performance with all three descriptors, which is also consistent with the result of Aisim.

\subsection{KITTI}
\subsubsection{Dataset and Implementation Details}
We use the KITTI dataset \cite{geiger2013vision} to evaluate the performance under real world. The KITTI dataset provides visual odometry consisting of 22 sequences. We split two trajectories from the Sequence 08 to simulate multi-robots global localization task. Additionally, the overlap of these two trajectories is about 70 meters. There is no Ground truth for segmentation and depth image for this sequence, therefore we need to generate these images manually. Here, we use \cite{zhu2019improving} and \cite{lee2019big} to generate these images. However, the quality of segmentation image is not ideal at the following aspects. First, the tree and grassland are merged into vegetation resulting in the system can not locate the correct geometry center. Additionally, the field of view angle of the visual sensor is small and it can not find big object's geometry center, like buildings. Last, there exists lost of repetitive scenarios which share the same objects like tree, car, building and pole. In this experiment, we will mainly focus on how our and previous methods perform under this challenging environment.

\begin{figure}[!t]
\centering
\subfloat[Original matches]{\includegraphics[width=1.5in]{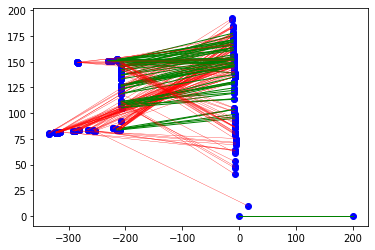}%
\label{fig_7_a}}
\subfloat[preliminary matches]{\includegraphics[width=1.5in]{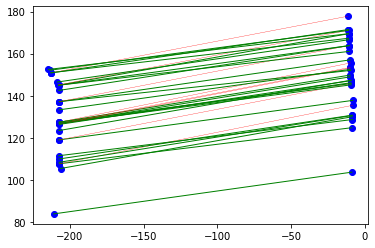}%
\label{fig_7_b}}
\hfil
\subfloat[Ori ($T_{error}=10m$ )]{\includegraphics[width=1.5in]{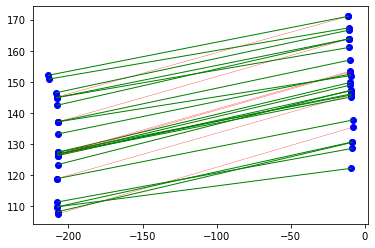}%
\label{fig_7_c}}
\subfloat[Ours ($T_{error}=10m$ )]{\includegraphics[width=1.5in]{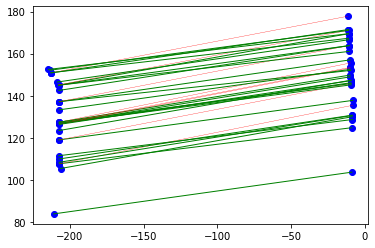}%
\label{fig_7_d}}
\hfil
\subfloat[Ori ($T_{error}=5m$ )]{\includegraphics[width=1.5in]{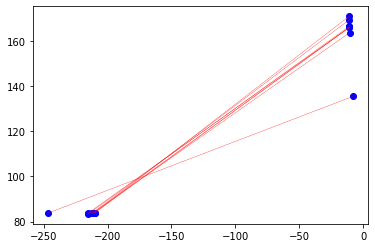}%
\label{fig_7_e}}
\subfloat[Ours ($T_{error}=5m$ )]{\includegraphics[width=1.5in]{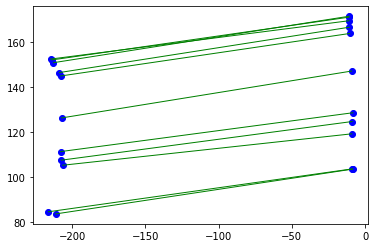}%
\label{fig_7_f}}
\caption{\textbf{A visualization comparisons on different RANSAC threshold.} This figure shows the main difference between our methods and original. For the previous method, the result (c) and (e) are directly from the original matches (a) via RANSAC. While ours method firstly get the preliminary matches (b) utilizing the neighbor constraints. Then get the result (d) and (f) from (b) via RANSAC method.}
\label{fig_7}
\end{figure}

Similarly to the previous approach, we employed two whole graphs for global localization and conduct a comparative analysis between our proposed method and previous. In this experiment, we quantify the precision of the final matches to evaluate the performance of our approach. The accuracy is determined based on the distance threshold of the matching points, where the distance less than 10m is considered as a correct match. The results are compared under the same RANSAC acceptable deviation/error threshold ($T_{error}$).
\subsubsection{Experimental Results and analysis}
We conduct comparative experiments with different RANSAC error threshold 10m and 5m and visualized the final results in Fig.\ref{fig_7} to better illustrate the advantages of our proposed method. In the $T_{error}=10m$ experiment, our method identified 26 out of 35 good matching points, whereas the original method only found 21 out of 28 matching points. Although the precision of both methods are similar, our approach is able to find more inliers, which is consistent with the results of previous experiments. This indicates that our method is more robust than the original method under semantically challenging environments.

We also observe that the RANSAC method may perform poorly when the threshold is set too low. In particular, when we set $T_{error}=5m$, there is a 44\% chance of interference from abnormal matches for the original method, resulting in completely wrong matches. On the contrary, our method rejects most of the abnormal matches during pre-processing. This is mainly due to the percentage of inliers in the sample is small and interfered by other abnormal matches. Our method effectively compensates for this gap by excluding most unreasonable outlier points during initial matching, significantly improving the proportion of inliers and avoiding unexpected results.

\section{Conclusion}

In this paper, we have identified that the proportion of overlapping parts between maps influence the performance of the system a lot. To address this issue, we have introduced a preliminary outlier rejection procedure into the pipeline. This strategic addition has led to substantial improvements in both the robustness and real-time performance of the system. 

Additionally, we conducted experiments on three different datasets to fully and fairly evaluate the performance of different approaches. Experiment results demonstrate that our approach outperforms previous works in multiple aspects, including precision, robustness, and processing time.

\addtolength{\textheight}{-4cm}   





\end{document}